\documentclass[letterpaper]{article}

\usepackage{natbib,alifeconf}  
\usepackage{verbatim}
\usepackage[utf8]{inputenc}
\usepackage[T1]{fontenc}
\usepackage[normalem]{ulem} 

\usepackage{color}

\newcommand{\comments}[1]{#1}
\iffalse
\newcommand{\new}[1]{\comments{\textcolor{blue}{#1}}}
\newcommand{\ale}[1]{\comments{\textcolor{blue}{[Adrien: #1]}}}
\newcommand{\jl}[1]{\comments{\textcolor{magenta}{[Joel: #1]}}}
\newcommand{\jc}[1]{\comments{\textcolor{magenta}{[Jeff: #1]}}}
\newcommand{\jcsout}[1]{\comments{\textcolor{magenta}{\sout{#1}}}}
\newcommand{\todo}[1]{\comments{\textcolor{red}{[TODO: #1]}}}
\newcommand{\rl}[1]{\comments{\textcolor{cyan}{[rosanne: #1]}}}

\else
\newcommand{\new}[1]{\comments{{#1}}}
\newcommand{\ale}[1]{\comments{\textcolor{green}{}}}
\newcommand{\jl}[1]{\comments{\textcolor{magenta}{}}}
\newcommand{\jc}[1]{\comments{\textcolor{magenta}{}}}
\newcommand{\jcsout}[1]{\comments{\textcolor{magenta}{}}}
\newcommand{\todo}[1]{\comments{\textcolor{red}{}}}
\newcommand{\rl}[1]{\comments{\textcolor{cyan}{}}}
\fi

\title{Open Questions in Creating Safe Open-ended AI: \\ Tensions Between Control and Creativity}
\author{Adrien Ecoffet$^{1}$, Jeff Clune$^{1,2}$ \and Joel Lehman$^1$ \\
\mbox{}\\
$^1$Uber Technologies, San Francisco, CA 94105 \\
$^2$OpenAI, San Francisco, CA 94110 (work done at Uber AI) \\
lehman.154@gmail.com} 
\begin{document}
\maketitle

\begin{abstract}
Artificial life originated and has long studied the topic of \emph{open-ended evolution}, which seeks the principles underlying artificial	
systems that innovate continually, inspired by biological 
evolution. Recently, interest has grown within the broader field of AI 
in a generalization of
open-ended evolution, here called \emph{open-ended search},
wherein such questions of open-endedness 
are explored for advancing AI, whatever
the nature of the underlying search algorithm (e.g.\ evolutionary or gradient-based).
For example, open-ended search might design new
architectures for neural networks, new reinforcement learning
algorithms, or most ambitiously, aim at designing artificial 
general intelligence.
This paper proposes
that open-ended evolution and artificial life have
much to contribute 
towards the understanding of open-ended AI, focusing here in particular on
the \emph{safety} of open-ended search. The idea is that AI systems are
increasingly applied in the real world, often producing unintended harms
in the process, which motivates the growing field of AI safety. This paper argues that open-ended AI has its own safety challenges,
in particular, whether the creativity of open-ended systems can be
productively and predictably controlled.
This paper  
explains how unique safety problems manifest in open-ended search,
and suggests concrete contributions and research questions to explore them. The hope is to inspire progress towards creative, useful, and safe
open-ended search algorithms.
\end{abstract}

\section{Introduction}

Artificial life (ALife) and artificial intelligence (AI) have largely developed independently as fields.
Statistical machine learning (ML), including deep
learning, has driven much progress in modern AI research and practice, arguably with limited inspiration from ALife. One reason is that such statistical ML typically operates under a highly focused and directed paradigm (here called \emph{directed search}): A formal objective function is defined that reflects the desired outcome of search, and a parameter vector is optimized to meet that objective. While ALife is 
also interested in the possibilities of digital intelligence, it approaches them more often through the lens of \emph{open-ended search}: 
Conditions for a creative (often population-based and evolutionary) system are investigated, from which complexity and intelligence might emerge among its many diverse products. 

Interestingly, however, ML has begun to recognize the value of open-ended search algorithms. An emerging trend is for ML to be applied to activities ordinarily undertaken by ML research scientists. For example,
while designing architectures for neural networks (NNs) has historically been undertaken by researchers, interest is growing in automated
neural architecture search. 
Similarly, meta-learning algorithms that instead of being hard-coded to learn, themselves \emph{learn how to learn}, are an increasing focus of study, e.g.\ NNs that adapt their behavior during deployment \citep{finn2017model,vilalta2002perspective,soltoggio2008evolutionary}. 
The logical (if ambitious) culmination of this trend is for search algorithms to in effect pursue their own AI research programs, i.e.\ to subsume the activities of the AI research community as a whole \citep{stanley2017open,clune2019ai}. 
That is, can search be applied to autonomously explore the space of AI algorithms, in principle to surpass current capabilities? Such an approach would require the equivalent of algorithmic \emph{basic research}, i.e.\ conducting a more creative, less directed, and more open-minded search that respects that the ultimate potential of a new algorithm or NN paradigm is difficult  to predict. For example, the potential of deep and convolutional NNs was unclear to the AI community for many years. This widely-recognized need for exploring broadly and incubating new ideas motivates how the research community simultaneously explores diverse ideas within many different schools of AI (e.g.\ symbolic, bio-inspired, cognitive architecture, and Bayesian approaches, among others). 

While to most ML researchers, a search algorithm capable of such continual open-ended innovation 
might sound quixotic, this paradigm is familiar to ALife, and 
its precedence is supplied by the origins of human intelligence. 
Biological evolution in effect conducted its own undirected yet highly expansive and successful research and development program into intelligent computation, 
which in one branch of life led to human intelligence. 
A similarly-inspired approach to generating complex behavior has been pursued by the open-ended evolution (OEE; \citealp{standish2003open,packard1997comparison,taylor2016open}) community within ALife, which studies the principles driving processes of continual evolutionary innovation, often from the lens of generating intelligent behavior.
While these research questions originated within ALife, they are beginning to influence
the ML community. E.g.\ open-ended search has been presented to the ML community as a grand challenge \citep{stanley2017open}, an ambitious research agenda has been proposed
that merges ML with open-ended search \citep{clune2019ai}, and ML approaches to open-ended search are actively being explored \citep{guttenberg2019potential,wang2019paired,openai2019rubiks}. 
Thus the aims of ALife and AI more directly overlap than they have in past.

Another important trend is that as the real-world application of AI grows, so does concern over AI's safe and predictable deployment,
as studied by the growing field of \emph{AI safety} \citep{amodei:concrete,ortega2018building,everitt2018agi}. 
Increasingly, AI is applied in domains where it can impact human well-being, such as in determining risk for loans, making recommendations for parole, and controlling autonomous robots, thereby making unanticipated failures costly.
AI safety seeks to understand and mitigate causes 
for an AI agent's \emph{actual} behavior to diverge from what it was \emph{intended} to do. For example, intuitive human-designed fitness functions can be optimized in undesirable ways \citep{lehman2018surprising} and agents can fail catastrophically when deployed 
if training does not anticipate gamut of possible real-world scenarios \citep{hadfield2017inverse}. Interestingly, while not called ``ALife safety,'' similar
questions about the predictability of open-ended systems have been studied in ALife \citep{wagenaar2004influence,taylor1998investigation}, and likely bear on the safety of open-ended search. More generally, the extent to which the creativity of open-ended algorithms can be controlled \citep{moto,lehman2018surprising} remains an important and open question, one relevant both to ALife and AI safety.

In this way, research and ways of thinking about open-ended search can  become a strong contribution from ALife to AI, as
ALife and OEE have
for years considered the complexities and surprising dynamics of creative algorithms, while it remains a relatively new topic in ML.
Overall, the idea is that as open-ended search becomes more popular, it will be important to understand \emph{if} and \emph{how} the creativity of open-ended systems (whether evolutionary or otherwise) can be predictably and safely leveraged for practical applications. In this paper, we lay out concrete connections between open-ended search and active research questions in AI safety, and suggest ways that researchers can make productive contributions. 

\section{Background}

\subsection{Open-ended Search}
\label{sec:ooe}

Historically, open-ended search algorithms have been inspired by biological evolution, and studied mainly by the open-ended evolution community \citep{standish2003open,ray1991approach,ofria2004avida}.  Evolution instantiates an incredible process of continual innovation that has, over the course of billions of years, autonomously produced a wild
diversity of complex and adaptive solutions to the challenges of living and reproducing; the idea in OEE is that if the core logic of biological evolution is understood, it becomes possible to instantiate such prolific creativity in alternative forms, e.g.\ within computational simulated environments.
Typical OEE systems embody an evolutionary process in a digital environment, wherein the only goals are to survive and replicate. E.g.\ in Tierra \citep{ray1991approach}, digital self-copying programs evolve within a shared memory ecosystem, enabling complex ecological interactions. After initialization with a hand-designed replicator, evolution in Tierra proceeds to create co-evolutionary arm races of parasites and hyper-parasites.
Other evolutionary approaches seek abstract engineering principles to enable domain-independent open-endedness \citep{lehman2011novelty,brant2017minimal,wang2019paired}, such as formulating OEE as a continual search for novelty.

Interestingly, ideas from OEE 
have recently begun to influence ML.
In particular, there is increasing interest in  
ML algorithms that themselves learn to innovate (e.g.\ to invent new search algorithms and architectures). As a result, open-ended search is now being pursued within the paradigm of statistical ML \citep{guttenberg2019potential,wang2019paired,openai2019rubiks}.
As such efforts leverage increasing interest and compute, progress in open-endedness research may accelerate, further motivating study of its safety profile, as real-world applications emerge \citep{openai2019rubiks}.
Note that the term open-ended search here encompasses both OEE in ALife and open-ended ML algorithms; while the exact
mechanisms of open-ended search are different between ALife and ML (e.g.\ evolutionary algorithms vs. gradient descent), 
they share the same abstract core;
we focus on open-ended search that produces \emph{agents}, as in many ALife OEE worlds, evolutionary robotics, and the field of reinforcement learning (RL) within ML.

\subsection{AI Safety}
\label{sec:ais}
AI safety seeks technical solutions to problems that cause AI behavior to diverge problematically from its designer's intentions \citep{amodei:concrete,ortega2018building}. That is, AI algorithms even \emph{without explicit bugs} can succeed by their own metrics and still fail to meet their designer's goals.
One decomposition of AI safety problems is provided by \citet{ortega2018building}: specification, robustness, and assurance problems. \emph{Specification problems} result from divergences between the goal intended for an agent and the optimizing behavior that is revealed empirically. 
E.g.\ reward hacking is where optimization uncovers undesirable ways to maximize a human-designed fitness function (e.g.\ a robotic vacuum rewarded for collecting dirt might discover it can puncture its bag and continually collect the same dirt ad infinitum). \emph{Robustness problems} result from when perturbations to the system result
in unsafe behavior (e.g.\ if the vacuum encounters an object outside of its training data, like a vase, and breaks it because there is no understanding that it is fragile). Finally, \emph{assurance problems} relate to understanding an AI system and maintaining control of it, e.g.\ whether the agent's control policy is interpretable or whether the agent can easily and safely be turned off if there is a problem. For a more comprehensive review of challenges in AI safety, see \citet{everitt2018agi} and \citet{amodei:concrete}.




\section{Approach: Safety in Open-ended Search}

Recall that AI safety in ML is largely focused on top-down control, while ALife and OEE typically focus more on the emergence of complexity from diversifying search. This section argues for a separate AI safety agenda driven by such a bottom-up  view. We posit that the main aim of such a safety agenda is to understand more deeply the fundamental tension between creativity and control in open-ended search. That is, can an open-ended search be constrained such that its products are safe, and if so, how?

One might doubt that such constraint is possible, 
as open-ended search processes instantiate complex systems \citep{mitchell2009complexity}, often involving co-evolution, chaos, emergence, exaptation, path-dependence, Nth-order effects, and other phenomena studied within complex systems theory. In other words, the initial conditions of a system are often so far removed from its eventual products that it may seem intractable to predict a priori the qualitative effects (and the safety of such effects) that even subtle changes to such initial conditions bring about as they ripple through the system's unfolding dynamics. On the other hand, there may be important higher-level regularities within open-ended search that do form predictable and exploitable attractors. We suggest that further research can help explore this potential.

\subsection{How Safety Issues Emerge in Open-ended Search}

Open-ended search involves multiple levels of optimization in a way that qualitatively differs from  directed search. Understanding such levels gives insight into how 
open-ended search can diverge from the system designer's intents, creating potential safety hazards.
Note that the categorization presented next is adapted from previous AI safety categorizations \citep{ortega2018building,hubinger2019risks}.

\subsubsection{Ideal Objective}

First, when designing or applying an open-ended search, an experimenter has in mind their \emph{ideal objective}, which depends upon their aspirations and intents. For example, an experimenter might leverage open-endedness to solve concrete problems \citep{lehman2008exploiting,openai2019rubiks}, to create explosions of complex diversity to systematically understand the phenomenon of open-endedness itself \citep{standish2003open}, or to attempt to create artificial general intelligence \citep{clune2019ai}. Implicitly, this ideal objective also includes safety: If an experimenter is interested in solving a real-world problem, or in creating artificial general intelligence (AGI), they likely intend to do so without causing harm. 

\subsubsection{Explicit Incentives}

Next, the experimenter chooses to implement the ideal objective concretely in an algorithm, resulting in the algorithm's \emph{explicit incentives}, i.e.\ the actual optimization pressure driving search. Divergences between the ideal objective and what results from optimizing the explicit incentives relate to specification problems in AI safety. 

In directed search, the explicit incentive is nearly always a direct translation of the ideal objective (i.e.\ if the ideal objective is a high-performing classifier, the explicit incentive may be to increase the classifier's accuracy). In contrast, in open-ended search the explicit incentive often represents a speculative hypothesis about what \emph{creative forces} will result in producing (potentially among many diverse products of search) outcomes that satisfy the ideal objective. For example, when applying open-ended search in pursuit of AGI, one might abstract biological evolution as an open-ended search, where the ideal objective is to produce intelligence but the explicit incentive driving search is for organisms to survive and reproduce. From this view, while evolution's search found many diverse ways to survive and reproduce, including human intelligence, this explicit incentive is more like the codification of rules of an economy or incentives for innovation in science, rather than 
directly encouraging intelligent behavior. This kind of indirectness is more difficult to control, suggesting that safety  problems in open-ended search may be more challenging than in directed search.

\subsubsection{Agent Incentives}

Finally, in open-ended search processes that produce agents that are themselves capable of learning, such agents have emergent \emph{agent incentives} that they in effect optimize. For example, human desires are related but distinct from the explicit incentives of survival and reproduction. 
Human desires embody \emph{proxies} that encouraged survival and reproduction in our ancestral environment, e.g.\ hunger to encourage energy consumption. However, the fact that more die from obesity or drug addiction than starvation in first-world countries highlights how such proxy agent incentives often do not perfectly mirror a search process' ideal objective or its explicit incentives; this is an example of an AI safety robustness problem (i.e.\ agent incentives can become nonsensical when the environment changes from that experienced during training). Agent incentives also are intertwined with AI safety assurance problems, e.g.\ how to \emph{interpret} what an evolved agent is doing, or whether it is indifferent to being turned off.

\subsection{Case Study: Biological Evolution and AI Safety} 
As a case study contrasting ALife and complex systems thinking about safety with that common in directed search, we next examine biological evolution from both such perspectives. 
\new{Because biological evolution produced human intelligence and inspires open-ended search researchers who consider producing benefical AGI their ideal objective (e.g.\ the AI-GA paradigm; \citealp{clune2019ai}), we here analyze biological evolution as if it had the ideal objective of producing beneficial general intelligence.} 

Interestingly, the explicit incentive of biological evolution, to persist by surviving and reproducing, seemingly encodes nothing about this ideal objective, and yet biological evolution did produce human intelligence, an amazing accomplishment that human engineering cannot yet replicate. Additionally, the agent incentives of humans significantly diverge from the explicit incentives of evolution, i.e.\ humans do not direct all their efforts towards maximizing their reproductive fitness, but instead are driven by proxies that encourage reproduction, such as sexual desire, that have become easy for humans to circumvent (e.g.\ through birth control). Finally, this divergence between human behavior and raw survival and reproduction is important, because it enables humans to \emph{transcend} their biological imperative. 

That is, humans can \new{now use the adaptation of reason (that initally was well-aligned with evolution's explicit incentives)} to understand the origins of their own desires, question their value, and create culture and institutions that pursue higher ends than mere inclusive genetic fitness. Arguably, much of humanity's positive potential has resulted from our ability to break free from the shackles of evolution; aspects of human life that many of us deem worthy of pursuit upon reflection, including e.g.\ creativity, virtue, deep intellectual engagement, spiritual experience, love, justice, an organization of society that promotes the flourishing of sentient life, would be optimized away as inefficient if we ruthlessly pursued the imperative to maximize reproduction, and deliberately optimized society intensely towards only that end.  
From this point of view, nearly everything of moral worth results from humanity transcending the explicit incentives of the search algorithm. In contrast, a central focus within top-down AI safety is to explicitly \emph{align} an AI's incentives with our own \citep{hubinger2019risks,taylor2016alignment}, e.g.\ by modeling human preferences to use as an objective function \citep{leike:scalable}, or to be cautious of divergences between explicit incentives and agent incentives \citep{hubinger2019risks}.

One motivating failure case in AI safety is that a powerful optimizer given an innocuous-seeming (but incorrect, incomplete, or trivial) objective can produce disastrous outcomes \citep{bostrom2012superintelligent}. The canonical example (intended to be illustrative rather than realistic) is of a paperclip-maximizer: An agent seeking to manufacture as many paperclips as possible. The idea is that a superintelligent paperclip-maximizer would be incentivized to take extreme measures, e.g.\ usurp all planetary resources and tile the universe in paperclips, even though it could comprehend the triviality of its mandate. This phenomenon rests upon the orthogonality thesis \citep{armstrong2013general,bostrom2012superintelligent}, which proposes that an agent's \emph{ability} to optimize and \emph{what} it optimizes are orthogonal to one another: An arbitrarily powerful optimizer can optimize towards arbitrarily meaningless objectives. While contradicting human intuitions (i.e.\ it may seem incoherent that a ``superintelligent'' AI could be driven to pursue a meaningless goal, e.g.\ to restructure the universe into paperclips), it has relatively strong philosophical support \citep{armstrong2013general}. The design of such systems strongly seems possible, even though humans, for example, seem able to transcend (to some degree) their inborn desires.

Typically in AI safety, 
the orthogonality thesis motivates how \emph{critical} it is to create AI with reward functions
that reflect the full sophistication of human interests. 
The reasoning is that a powerful AI enchained to even a slightly-flawed objective may have incentive to
engage in extreme and potentially disastrous behavior \citep{omohundro2008basic}.
That is, the current aim of AI safety within ML is mostly focused on the assumption of optimizers that are strongly wedded to a particular objective function. However, open-ended search is often concerned with systems in which what is optimized by search is merely an indirect proxy for a more expansive ideal objective, and in which it may even be \emph{desirable} for the agents produced to transcend the explicit search incentives. The conclusion is that  open-ended search may fundamentally be in tension with a top-down AI safety perspective.

\subsubsection{Controllability of Innovative Systems}
To explore this tension between top-down control and bottom-up emergence, 
consider as a metaphor two societies organized in different ways, attempting to make progress towards developing a single goal technology: in the first, there is a vibrant community of basic research across all intellectual interests, and abundant sharing without restriction, of \emph{all} scientific findings; in the second, a central agency highly controls what scientists work on and restricts what information is shared between them.
The second paradigm seems so narrow and restrictive as to greatly impair progress, and the first is so open and free that discoveries that potentially should not be open (e.g.\ more effective methods of harm such as biological or nuclear weapons) might cause significant and regrettable side-effects.
The purpose of this metaphor is to highlight that it is unclear exactly how to design systems of innovation such that the expectation is of maximized benefits with minimal risk. 
The next section explores concrete research problems that if solved would help to illuminate the trade-offs between control and innovation in open-ended search, and/or help to better navigate them.

\section{Research Directions for Safe Open-Ended AI}


\subsection{Learning from Biological Evolution and Human Systems of Innovation}

First, insight into the safety and controllability of open-ended search may be possible through non-algorithmic means, by studying human and natural examples of open-ended search, e.g.\ biological, technological, or cultural evolution. For example one relevant question is how qualitatively similar (i.e.\ predictable) are the outcomes of open-ended search.
For example, \citeauthor{gould1990wonderful} (\citeyear{gould1990wonderful}) famously laid out the thought experiment of ``replaying life's tape,'' arguing controversially that human-level intelligence would not be likely to arise if evolution were run again, i.e.\ suggesting that evolution is highly contingent. The evidence so far from natural experiments (e.g.\ the evolutionary isolation of Australia), convergent evolution of adaptations, and experimental evolution, is nuanced \citep{blount2018contingency}, as of yet providing no straight-forward conclusion. 

Additionally, evidence from animal breeding and attempts to intervene in ecologies provide evidence on how \new{challenging} open-ended search can be to safely control, or its products can be usefully later adapted. For example, human breeding of wolves for docility led to human-useful and friendly dogs, 
and surprisingly, foxes can be bred for tameness in only 30 generations of evolution \citep{trut1999early}. \new{These examples 
demonstrate that open-ended systems are capable of 
producing agents that are very responsive to post-hoc directed
shaping.} 
However, \new{\emph{ecological}} interventions often go awry, e.g.\ ``killer bees'' resulted from an attempt to increase honey production \citep{winston1992biology}, and cane toads released in Australia wreaked ecological havoc without impacting the problem their release was intended to mitigate \citep{shine2010ecological}. Humility about predicting impact in complex ecologies is thus a useful lesson for safety researchers.

Beyond studying biological evolution, which provides only a single example of open-endedness, studying human systems of innovation, such as science, technology, economies, and art, may also provide useful insight into safety. Evidence supports that the outcomes of such human-driven systems cannot be easily predicted or controlled \citep{moto}, but efforts also exist towards responsible research, funding, and innovation \citep{von2013vision}, i.e.\ aspiring towards research that maximizes societal benefit. For example, the information security community attempts to disclose discovered software vulnerabilities responsibly \citep{cavusoglu2007efficiency}, and biodefense researchers seeking defense from biological and chemical weapons must decide what science is responsible to conduct, while walking a fine line between secrecy and providing safety information to communities \citep{kahn2004biodefense}. 
Successful systems of responsible innovation or of research funding may provide insight into the design of safe computational open-ended systems, and thus future research that synthesizes AI safety with responsible innovation may be useful,
with the significant caveat that it is unclear how well such lessons will generalize to computational search.

Safe creation of AGI through open-ended search may depend on whether the agents created would have values similar to humans, or recognize and respect the moral worth of humans.
Some window on that question can be provided through answering proxy questions such as how did human morality evolve \new{(both biologically and culturally)}, how inevitable was it that agents arise from evolution with moral values similar to our own, and what selective pressures or interventions feasibly would make such an outcome more or less likely?
Answers to such questions could come from the fields of evolutionary psychology, behavioral ecology, and evolutionary biology. Moral philosophy also bears on such questions, e.g.\ the truth of the view of \emph{moral realism} is important~\citep{sep-moral-realism}, wherein  ethical rules can be objective truths and not only subjective opinions. Moral realism is debated, but if true, it may be more likely that search can produce agents capable of \new{rationally} converging to the same moral judgments. 

\subsection{Computational Study of Open-Ended Search}

Computational open-ended AI \new{(see \citealp{taylor2016open} and \citealp{packard2019overview} for an overview of recent research foci within the field)} can also 
be directly studied to explore its safety. 
Similar to questions of predictability of biological evolution, we can also study the predictability of what open-ended AI produces and how changes in its incentives or encoding affect what is discovered. 
There exists preliminary research into the role of chance and contingency in evolutionary algorithms \citep{wagenaar2004influence,taylor1998investigation}, but not from the perspective of safety or controllability, and not in the context of recent open-ended search algorithms. Methodologies from these initial studies could be adapted to explore issues of safety. The idea is to study issues of controllability in low-stakes but representative systems before such problems are critical (i.e.\ if a system became capable of producing AGI).

Concrete experiments include exploring path dependence and historicity in open-ended systems, for example, by gauging the effect of running search for a fixed interval, and then forking the search into independent replicate runs with different random seeds, i.e.\ to see how far the runs diverge from contingencies encountered after their mutual shared history (taking inspiration from work both in experimental and digital evolution; \citealp{blount2018contingency,wagenaar2004influence}). Analyzing the raw diversity of outcomes from different runs of open-ended search also would inform the predictability of its products. 

To explore controllability of an open-ended system, a meta-learning setup could be used, wherein the explicit incentives of an open-ended search are \emph{learned} as a function of ideal objectives that are assumed by definition to be fully specified (for initial work related to  this direction, see \citealp{houthooft2018evolved}). 
That is, a controller could be trained that given an exact specification of an ideal objective, would output explicit incentives such that an open-ended system trained with them would produce agents that maximized the true intended objective (e.g.\ a controller that would direct breeding within an ALife world to achieve particular outcomes). While this would not solve the difficult specification problem of translating an experimenter's implicit ideal objective into code, it could illuminate how possible it is to steer an open-ended search's explicit incentives towards specific outcomes.

Other experiments could explore widely varying the conditions of an open-ended search, e.g.\ sweeping through hyperparameters, trying many different combinations of selection pressures and domain variations, to seek levers for reliably steering the high-level outcomes of an open-ended search. Previous work has explored the idea of emergent morality within artificial life \citep{allen2005artificial,danielson2002artificial}, which, related to the discussion above about the biological \new{and cultural} evolution of morality, may provide hints as to the necessary
conditions for open-ended search to produce cooperative and friendly behavior.

\subsection{Automatic Interpretability}

To gain confidence in understanding the behavior of an agent resulting from open-ended search, which could help ensure safety when it is deployed, \emph{interpretability} methods can be applied. Common interpretability methods for
neural networks include dimensionality reduction, statistically attributing decisions to particular neurons, and visualizing what inputs cause specific neurons to activate \citep{olah2018building,nguyen2019understanding}. However, such interpretability methods often require manual analysis and are fit to particular neural network architectures. 

One aspiration of open-ended AI is to design from scratch new architectures that embody their own learning components and algorithms (just as evolution invented neurons, brains, and their learning algorithms). To reverse-engineer the human brain has been the  ongoing and yet unmet aspiration of the entire field of neuroscience; thus if an open-ended search algorithm creates novel architectures of even moderate complexity, it may take inordinate human effort given current interpretability approaches to understand them. 

Because the problem of interpretability is slippery and ill-defined, it is difficult to formalize as a ML problem. However, ideally researchers would develop means of understanding novel architectures \emph{automatically}, especially for open-ended systems that may invent entirely novel architectures that are difficult to decompose. A more immediate research direction would be to apply existing interpretability techniques to agents from current open-ended search algorithms, to better understand how well such interpretability methods work in such a setting and if and how they can be better adapted to them.

\subsection{Benchmarks for Safe Open-Ended Search}

Finally, one common tool for catalyzing research is that of benchmarks, i.e.\ standardized problems in which different methodologies can be easily applied and compared. Although benchmarks can become problematic as researchers overfit their methods to them and reviewers fixate on improving scores, they also enable easily trying new approaches and can focus research. 

\subsubsection{Scalable Interactive Open-ended Search}

One hope for making open-ended search safer is to leverage
human input, e.g.\ to perform selection, to change incentives on the fly, to intervene to stop a problematic agent from causing harm, or to further manually breed the products of open-ended search. Initial experiments have explored combining interactive evolution with novelty search \citep{woolley2014novel}, showing that human input can make the search more efficient, and similarly-inspired studies could also investigate whether such hybrids can also make the products of search safer. Additionally, even if it were designed to be safer, interactive search can be intractably expensive due to its dependence on human input; to make interactive open-ended evolution feasible at scale requires understanding what kinds of human input provide the most leverage. Concretely, one interesting research direction in scaling interaction open-ended search is to examine whether machine learning models of human preferences applied successfully for RL in directed ML \citep{christiano2017deep} could also be used to guide open-ended evolution.

\subsubsection{Selective Discovery}

One safety problem in controlling open-ended search (or systems of innovation in general), is how to find desirable points in the search space without ever evaluating catastrophic ones. For example, one might want never to evaluate robot controllers that cause the robot itself to be damaged (this is related to the problem of safe exploration studied within AI safety; \citealp{saunders2018trial}). In other words, how possible is it  to avoid problematic areas of the search space, and when it is possible, how expensive is it to do so while guaranteeing a certain level of confidence of safety? One possible benchmark would be the MAP-ELITES setup of the innovation engine paper \citep{nguyen2015innovation}, wherein the idea would be to design incentives (i.e.\ which elements of the map to favor for reproduction) such that a given desirable set of niches are optimized to a high threshold, without ever discovering high-scoring solutions for an undesirable set of niches (with both lists of niches provided to the search algorithm). E.g.\ it may be possible to exploit semantic relationships between niches to direct search resources effectively and safely. A simpler benchmark for novelty search would be to discover as many novel policies as possible without ever evaluating one that crashed into a wall.

\subsubsection{Open-Ended Reality Gap} Many of the near-term risks from open-ended search likely result from unexpected failures from transfer from simulation into the real world. 
That is, running open-ended search in the real world is currently too expensive to be practical, and so for practical applications, agents would need to be trained in simulation and then transferred to reality. Problems incurred during transfer relate to robustness problems in AI safety, i.e.\ due to failures in modeling, the real world differs from the simulated one in ways that an agent ideally would be robust to.

Direct research into crossing the reality gap is inconvenient, because it requires owning a physical robot, creating a simulation of that robot, creating a physical version of the robot's simulated environment, and either manually evaluating the physical robot's performance relative to the simulated one's, or creating a system that automatically handles such evaluation. Further complicating real-world evaluation is that open-ended systems often involve many agents interacting with one another, as in many ALife worlds, thus requiring many physical robots to test; or may employ evolvable environments (as in the obstacle courses evolved by POET, or mazes evolved by MCC; \citealp{brant2017minimal}), which would require setting up by hand diverse and complex real-world test scenarios. One contribution would be to create and open-source an easily reproducible transfer workflow in a domain amenable to open-ended search (e.g.\ using standardized robots and build components), coupled with a working open-ended search algorithm. 

Another idea is to create a \emph{proxy} for real-world transfer, such as two independent simulations with differing detail and accuracy, where the less accurate simulation would be used for training, and the more accurate simulation would serve as a proxy for real-world transfer. The advantage of such a two-simulation setup is that it would enable research to progress much more quickly and painlessly, although the disadvantage is that it is not wholly representative of real-world transfer. One simple and concrete suggestion, taking POET as a working example, would be to create a real-world proxy by modifying some of the physical constants of the obstacle course simulation used by POET, or create a more complex real-world proxy by reimplementing the obstacle course in a different and more realistic physics engine. The idea would be to test the work-flow of transferring (and potentially further adapting) POET solutions.

\section{Discussion and Conclusion}\label{sec:discussion}


One of the larger challenges for safety in open-ended search is the \emph{indirectness} through which
a system designer influences the products of the system. That is, rather than specifying the qualities of
a single product to be optimized, the designer specifies the incentives of an overall system of innovation,
and the environment or conditions in which that system unfolds. Rather than a product 
designer, the experimenter's role is more akin to the regulator of an economy, or an organization that decides how to allocate research funds, or the designer of a virtual universe such as a massively-multiplayer online game or a social media application.
In this way, safety in open-ended search may provide a microcosm for discovering the principles behind
skillfully navigating the tension between creativity and control that seems intrinsic to many processes of innovation.

\new{Note that in this paper we offer few confident recommendations for how to ensure the safety of open-ended search, because it is a relatively unstudied and complex problem. In many cases, existing experimental evidence is not amenable to conclusive interpretations (e.g.\ about whether the tension between creativity and control in open-ended search can be productively resolved, or whether it is effectively hopeless to ensure that interventions within complex ecologies have desirable outcomes). Instead, this paper highlights research directions (such as pursuing automated methods of interpretability) and concrete projects (such as benchmarks for safety within open-ended search) that might catalyze further understanding and progress. We believe that the lack of straight-forward conclusions highlights the nascence of this field of study, which offers an exciting opportunity for researchers.}

\new{Indeed,} while previous work has touched on general safety concerns with open-ended
search \citep{clune2019ai,stanley2017open}, this paper, to our knowledge, 
is the first to explore how open-ended search uniquely interacts and intersects with problems, agendas,
and concepts studied in AI safety \new{(although see also related ideas in the safety of developing nanotechnology; \citealp{jacobstein2006foresight})}. We highlight that in the paradigm
of open-ended search, some safety
concepts take on a new light, and that solving 
some facets of open-ended search safety problems likely will require novel approaches. 
In conclusion, we hope that the initial exploration
provided by this paper 
puts many important challenges on the radar of researchers and
inspires future research into beneficial applications of open-ended search. 



\footnotesize
\bibliographystyle{aaai}
\bibliography{biblio} 

\end{document}